\documentclass[10pt,twocolumn]{article}

\usepackage[utf8]{inputenc} 
\usepackage[T1]{fontenc}    
\usepackage[expansion=false]{microtype}      
\usepackage{amsmath,amssymb}
\usepackage{booktabs}
\usepackage{graphicx}
\PassOptionsToPackage{hyphens,obeyspaces,spaces}{url}
\usepackage{hyperref}
\usepackage{xurl}
\usepackage[numbers]{natbib}
\hypersetup{
    colorlinks=true,
    linkcolor=blue,
    citecolor=blue,
    urlcolor=blue,
    pdftitle={Tiny Recursive Models on {ARC-AGI-1}: Inductive Biases, Identity Conditioning, and Test-Time Compute},
    pdfauthor={Author Name},
}

\title{Tiny Recursive Models on {ARC-AGI-1}: Inductive Biases, Identity Conditioning, and Test-Time Compute}

\author{
  \textbf{Antonio Roye-Azar}\textsuperscript{1,2}, 
  \textbf{Santiago Vargas-Naranjo}\textsuperscript{1,2} \\[4pt]
  \textbf{Dhruv Ghai}\textsuperscript{1}, 
  \textbf{Nithin Balamurugan}\textsuperscript{1}, 
  \textbf{Rayan Amir}\textsuperscript{1} \\[6pt]
  \textsuperscript{1}Western University \quad
  \textsuperscript{2}Varonova Tech Inc.\\[4pt]
  \texttt{antonio@varonova.ca}\\
  \texttt{\{svargasn, dghai3, nbalamur, ramir2\}@uwo.ca}
}

\date{} 

\begin{document}

\twocolumn[
  \maketitle
  \begin{abstract}
  Tiny Recursive Models (TRM) were recently proposed as a parameter-efficient alternative to large language models (LLMs) for solving Abstraction and Reasoning Corpus (ARC) style tasks. The original work reports strong performance and suggests that recursive latent updates enable non-trivial reasoning, but it remains unclear how much of this performance stems from architecture, test-time compute, or task-specific priors. In this technical note, we empirically analyze the publicly released ARC Prize verification checkpoint for TRM on ARC-AGI-1. We report four behavioral findings and a small-scale efficiency comparison. First, we show that test-time augmentation and majority-vote ensembling account for a substantial fraction of reported performance: the 1000-sample voting pipeline improves Pass@1 by 10.75 percentage points over single-pass canonical inference. Second, a puzzle-identity ablation reveals strict dependence on task identifiers: replacing the correct puzzle ID with a blank or random token yields zero accuracy under the verification protocol. Third, a recursion trajectory analysis shows that most of the final accuracy is achieved at the first recursion step and that performance saturates after a small number of latent updates, indicating effectively shallow recursion. Fourth, early-stage training experiments under canonical versus heavy augmentation regimes suggest that heavy augmentation broadens the distribution of candidate solutions and improves multi-sample success before single-pass accuracy improves. Finally, we compare TRM with a naive QLoRA fine-tune of Llama 3 8B on canonical ARC-AGI-1, illustrating that TRM's non-autoregressive design achieves much higher throughput and substantially lower memory usage in this setting. Overall, our results suggest that TRM's strong performance on ARC-AGI-1 arises from an interaction between efficiency, task-specific conditioning, and aggressive test-time compute, rather than from arbitrarily deep internal reasoning dynamics.
  \end{abstract}
  \vspace{1em} 
]

\section{Introduction}

The Abstraction and Reasoning Corpus (ARC) was introduced as a benchmark for evaluating generalizable, human-like reasoning in artificial systems \citep{chollet2019measure}. ARC and its successors, ARC-AGI-1 and ARC-AGI-2, are explicitly designed to be difficult for standard large-scale pattern-matching methods, even when such methods succeed on more conventional vision and language tasks. Each ARC task consists of a small set of input--output grid pairs and requires the learner to infer and apply a transformation rule to new inputs.

Large language models (LLMs) have recently demonstrated non-trivial performance on ARC-style benchmarks when equipped with task-specific tokenization, chain-of-thought prompting, and test-time training or adaptation \citep{wei2022chain, snell2024scaling}. However, these systems typically have very high parameter counts and incur significant inference latency and memory costs, especially when they rely on multiple sampled trajectories or search-based decoding.

Tiny Recursive Models (TRM) were proposed as a contrasting approach that combines a small parameter footprint with recursive inference over latent states \citep{jolicoeurmartineau2025trm}. In the original study, a 7M-parameter TRM trained with heavy data augmentation and deep supervision reportedly achieves strong performance on ARC-AGI-1 and ARC-AGI-2 while operating at much lower memory and latency than typical LLMs. This has led to interest in TRM as a ``tiny but capable'' reasoning architecture.

Despite this interest, several aspects of TRM's behavior remain insufficiently characterized:

\begin{itemize}
    \item \textbf{Role of test-time compute.} The original paper uses extensive test-time augmentation and majority-vote ensembling, but the exact contribution of this ensemble, relative to the underlying single-pass model, is not fully quantified in a standardized setting.
    \item \textbf{Dependence on puzzle identity.} TRM includes puzzle-specific embeddings indexed by puzzle IDs. It is unclear to what extent performance depends on these identifiers, as opposed to the information present in the input grids alone.
    \item \textbf{Effective depth of recursion.} The architecture allows multiple recursive latent updates, and training uses deep supervision across many steps. It is not obvious how many steps materially contribute to the final prediction in a deployed checkpoint.
    \item \textbf{Impact of augmentation on the solution distribution.} Heavy augmentation is used both at training and test time, but its effect on accuracy and on the distribution of candidate solutions is not well understood.
\end{itemize}

Understanding these factors is important for at least two reasons. First, researchers may wish to know whether TRM's performance reflects generalizable reasoning or a combination of dataset-specific priors and test-time compute. Second, future work on recursive or tiny reasoning models can benefit from a clearer description of design trade-offs and failure modes.

This note does not propose a new architecture. Instead, it aims to provide a careful empirical characterization of an existing, publicly available TRM checkpoint, with a focus on the relative roles of identity conditioning, recursion depth, augmentation, and test-time compute.

\subsection{Contributions}

We analyze the ARC Prize verification checkpoint for TRM on ARC-AGI-1 and make the following contributions:

\begin{enumerate}
    \item \textbf{Ensemble contribution.} We reproduce the verification performance on ARC-AGI-1 and quantitatively separate the contribution of the 1000-sample test-time ensemble from that of single-pass canonical inference, showing that ensembling yields a substantial absolute gain in Pass@1 accuracy.
    \item \textbf{Puzzle-identity dependence.} We design a puzzle-identity ablation that replaces the correct puzzle ID with either a fixed ``blank'' identifier or random identifiers. Under both manipulations, accuracy collapses, indicating strict functional dependence on correct puzzle identity for this checkpoint.
    \item \textbf{Recursion trajectory analysis.} We evaluate intermediate outputs of the recursive inference process and compute accuracy at each recursion step. We find that most of the final performance is already present at the first step and that accuracy saturates within a small number of steps.
    \item \textbf{Training dynamics under augmentation.} We partially train TRM models under canonical-only versus heavily augmented regimes and compare early training behavior. Heavy augmentation broadens the solution distribution and improves multi-sample metrics before single-pass accuracy improves.
    \item \textbf{Efficiency and LLM baseline.} We compare TRM to a naive QLoRA fine-tune of Llama 3 8B on canonical ARC-AGI-1, measuring memory usage, throughput, and accuracy to illustrate TRM's efficiency and inductive bias advantages in this setting.
\end{enumerate}

Across all experiments, we use publicly available checkpoints, standard ARC-AGI-1 splits, and simple, well-documented metrics, and we explicitly discuss the limitations of our setup.

\section{Related Work}

\subsection{ARC and ARC-AGI Benchmarks}

ARC was proposed as a benchmark for general intelligence that emphasizes few-shot abstraction and compositional reasoning rather than large-scale supervised learning \citep{chollet2019measure}. ARC-AGI-1 and ARC-AGI-2 extend this setting and have been adopted in challenge settings and leaderboards, including recent ARC Prize competitions and public leaderboards \citep{arcprize2024arcagi1}. Public reports and leaderboards show that a range of systems, from program synthesis pipelines to frontier LLMs with specialized prompting and test-time training, achieve non-trivial but still subhuman accuracy, with a notable drop in performance when moving from ARC-AGI-1 to ARC-AGI-2 \citep{chollet2025arcagi2}.

These benchmarks are intentionally small and structurally diverse, which complicates standard large-scale training and evaluation practices. As a result, many systems rely on heavy data augmentation, hand-designed priors, or bespoke evaluation pipelines. The present work follows this tradition in focusing on ARC-AGI-1 as a concrete testbed, but differs in that we treat a single publicly released model (TRM) as an object of empirical analysis rather than proposing a new system to compete on aggregate scores.

\subsection{Recursive Reasoning Models and Tiny Recursive Models}

Hierarchical Reasoning Models (HRM) were proposed as a class of small recurrent architectures operating over latent states to solve structured puzzles such as Sudoku, mazes, and ARC-style tasks \citep{wang2025hrm}. HRM combines multiple recurrent modules operating at different ``frequencies'' with deep supervision across many steps and relies heavily on data augmentation. Follow-up analyses have suggested that deep supervision and augmentation contribute substantially to performance, and that the theoretical justification for some components may not strictly hold in practice \citep{arcprize2025hrmanalysis}.

Tiny Recursive Models (TRM) build directly on this line of work, simplifying the architecture while improving performance on several benchmarks \citep{jolicoeurmartineau2025trm}. TRM replaces the dual-network hierarchy with a single small network, makes the recursion mechanism more explicit, and incorporates architectural choices that favor efficiency on small grids. The original TRM paper reports strong performance on Sudoku Extreme, Maze Hard, ARC-AGI-1, and ARC-AGI-2, and emphasizes that a 7M-parameter model can outperform many larger systems when combined with heavy augmentation, deep supervision, and test-time ensembling.

Subsequent technical reports have examined TRM from complementary angles, including test-time adaptation and fine-tuning on new tasks \citep{mcgovern2025trelis}. These works generally treat TRM as a black box and focus on end-to-end performance. In contrast, our goal is to analyze the behavior of one fixed verification checkpoint and to isolate the roles of puzzle identity, recursion depth, and augmentation in that specific instance.

\subsection{Large Language Models, Chain-of-Thought, and Test-Time Compute}

Large language models have shown substantial progress on reasoning benchmarks, including ARC-like tasks, when equipped with chain-of-thought prompting and test-time adaptation. Chain-of-thought methods explicitly elicit intermediate reasoning steps, often improving performance at the cost of increased inference time \citep{wei2022chain}. More recent work formalizes the notion of test-time compute and shows that allocating additional sampling, search, or optimization budget at inference time can sometimes be as impactful as scaling model parameters \citep{snell2024scaling}.

Many of these methods rely on generating multiple candidate trajectories per query and aggregating them, for example via self-consistency or majority voting. This is conceptually similar to TRM's use of heavy augmentation and majority-vote ensembling, though the underlying architectures and compute regimes differ: autoregressive transformers have high token-level latency, while TRM predicts entire grids non-autoregressively.

Our work adopts this test-time compute perspective but applies it to a tiny non-autoregressive model. Instead of proposing a new inference scheme, we quantify how much of TRM's reported performance can be attributed to test-time compute, identity conditioning, and recursion depth within the existing verification pipeline.

\section{Experimental Setup}

In this section we describe our experimental setup in enough detail to permit reproduction. When we make specific choices, such as checkpoint selection or metric definitions, we motivate them and later discuss their limitations.

\subsection{Model and Checkpoint}

We study the ARC Prize verification checkpoint for TRM, which is publicly available via Hugging Face as \url{arcprize/trm_arc_prize_verification}. We chose this checkpoint for three reasons:

\begin{itemize}
    \item \textbf{Public availability.} Using a publicly hosted checkpoint allows others to reproduce our results without access to private training infrastructure.
    \item \textbf{Standard reference.} Documentation from the ARC Prize Foundation treats this checkpoint as the canonical reference for verifying TRM performance on ARC-AGI-1.
    \item \textbf{Fixed hyperparameters.} The checkpoint is trained with a fixed recursion depth of four latent cycles, whereas the original paper's strongest model uses six, providing a concrete, frozen target for analysis.
\end{itemize}

A limitation of this choice is that our results may not hold for all TRM configurations. We explicitly restrict our claims to this particular checkpoint and treat generalization to other TRM variants as future work.

\subsection{Dataset and Evaluation Split}

All main experiments are conducted on the ARC-AGI-1 public evaluation set, which consists of 400 tasks. Each task contains 2--3 training examples and 1--2 test examples. We use the same split and evaluation pipeline as the verification scripts associated with the TRM checkpoint.

We do not use ARC-AGI-2 or private competition splits in our experiments. Focusing on ARC-AGI-1 allows us to perform repeated ablations and per-step analyses on a manageable dataset, and it aligns with the evaluation setting for which the verification checkpoint was designed.

\subsection{Metrics}

We report accuracy as Pass@1, defined as the fraction of tasks for which the model's top-scoring output exactly matches the ground-truth output grid for the held-out test example(s).

For experiments involving multiple candidate outputs, such as test-time augmentation with multiple samples, we occasionally report Pass@1000 as a diagnostic metric, defined as the probability that at least one of the sampled outputs is correct. In the official TRM pipeline, performance is reported as Pass@1 for the aggregated majority-vote prediction, not as Pass@k over the raw sample set. We adhere to that standard for our main ensemble analysis and use Pass@k only when it serves an explanatory role.

\subsection{Augmentation Regimes}

TRM's training and evaluation pipelines both rely heavily on input augmentation. We consider two augmentation regimes:

\begin{itemize}
    \item \textbf{Canonical (aug 0).} Only the canonical ARC examples are used. No geometric or color augmentations are applied.
    \item \textbf{Heavy augmentation (aug 1000).} For each available demonstration pair, 1000 random augmentations are generated using color permutations, dihedral symmetry transformations, and translations within bounds that preserve task semantics.
\end{itemize}

For the existing TRM checkpoint, the heavy augmentation regime was used during training. For our training dynamics experiments, we train new models under each regime to partially disentangle the effects of augmentation.

\subsection{Computing Infrastructure}

All experiments that require a GPU use a single NVIDIA H100 with 80 GB of memory. For the main TRM ablations, this choice is not essential; TRM's memory footprint is low enough that smaller devices would suffice. The Llama 3 8B QLoRA baseline, however, benefits from the larger GPU memory.

When reporting efficiency metrics such as throughput and latency, we emphasize that these numbers are hardware-dependent and should be interpreted relatively (TRM versus Llama) rather than as absolute benchmarks.

\section{Experiment 1: Ensemble Contribution on ARC-AGI-1}

\subsection{Motivation}

The TRM verification pipeline uses 1000 augmented variations of each test puzzle and selects the final answer via majority voting. This procedure can substantially improve performance even for models that are individually weak, similar to how self-consistency and ensemble sampling can improve LLM accuracy \citep{wei2022chain, snell2024scaling}. It is therefore important to separate the contribution of the base model from the contribution of test-time compute.

\subsection{Design}

We evaluate the verification checkpoint on the ARC-AGI-1 public evaluation set under two conditions:

\begin{enumerate}
    \item \textbf{Paper mode (1000-sample ensemble).} For each puzzle, we generate 1000 augmented inputs using the same augmentation pipeline as in the verification code, run TRM on each augmented input, and aggregate the predictions via majority voting. We then compute Pass@1 for the resulting single prediction per puzzle.
    \item \textbf{Single augmentation mode.} For each puzzle, we present a single canonical input without augmentation, run TRM once, and directly compute Pass@1 from the single prediction. No voting is used.
\end{enumerate}

All other settings, including the checkpoint, recursion depth, and scoring function, match the verification scripts.

\subsection{Results}

As shown in Table~\ref{tab:ensemble}, the 1000-sample ensemble improves Pass@1 by 10.75 percentage points over single-pass canonical inference on ARC-AGI-1 (400 tasks).

\begin{table}[ht]
\centering
\caption{Impact of test-time ensembling on Pass@1 accuracy (ARC-AGI-1, 400 tasks).}
\label{tab:ensemble}
\resizebox{\columnwidth}{!}{%
\begin{tabular}{lccc}
\toprule
Evaluation mode & Augmentations & Voting & Pass@1 \\
\midrule
Paper mode (official)   & 1000 & Yes & 40.00\% \\
Single augmentation     & 1    & No  & 29.25\% \\
\bottomrule
\end{tabular}
}
\end{table}

\subsection{Interpretation}

This result confirms that test-time compute plays an important role in TRM's reported performance. At the same time, single-pass accuracy from a 7M-parameter model remains notable and motivates the deeper analyses that follow. We treat this experiment as a baseline that sets the scale of ensemble effects; later experiments on puzzle identity, recursion depth, and training dynamics should be interpreted against this backdrop.

\section{Experiment 2: Puzzle Identity Ablation}

\subsection{Motivation}

TRM includes a learned embedding table indexed by puzzle IDs. During training and evaluation, each puzzle is associated with a unique identifier that is mapped to a high-dimensional embedding and integrated into the model's input representation. While this design is visible in code, its behavioral consequences have not been systematically measured.

Conceptually, puzzle-ID embeddings could act as mild priors that group puzzles into families, or they could function as critical keys for retrieving task-specific behaviors. Distinguishing between these possibilities is important for understanding how TRM organizes its capacity across tasks.

\subsection{Design}

We perform a controlled inference-time ablation on the ARC-AGI-1 public evaluation set. For each puzzle, we consider three conditions:

\begin{enumerate}
    \item \textbf{Baseline.} Use the original puzzle identifier as in the verification script.
    \item \textbf{Blank ID.} Replace all puzzle IDs with a single fixed identifier whose embedding is trained but does not correspond to any evaluation puzzle.
    \item \textbf{Random ID.} Replace each puzzle ID with a random identifier that does not match the puzzle's true ID.
\end{enumerate}

In all conditions, we retain the full 1000-sample augmentation and majority-vote pipeline used in the verification setting. We then compute Pass@1 accuracy for each condition.

\subsection{Results}

As shown in Table~\ref{tab:puzzleid}, under both ablated conditions (blank ID and random IDs), accuracy collapses to zero, despite the fact that the input grids, augmentation pipeline, and network parameters remain unchanged.

\begin{table}[ht]
\centering
\caption{Effect of puzzle identity perturbations on Pass@1 accuracy (ARC-AGI-1, 400 tasks).}
\label{tab:puzzleid}
\resizebox{\columnwidth}{!}{%
\begin{tabular}{lcc}
\toprule
Condition & Puzzle ID input & Pass@1 \\
\midrule
Baseline          & Correct IDs            & 40.00\% \\
Blank ID          & Fixed ``blank'' token  & 0.00\%  \\
Randomized IDs    & Random token per task  & 0.00\%  \\
\bottomrule
\end{tabular}
}
\end{table}

\subsection{Interpretation}

These results support the conclusion that the verification checkpoint is functionally dependent on correct puzzle identity. The puzzle ID does not behave like a weak prior that can be removed with only a moderate performance drop; instead, it is necessary for producing correct outputs on ARC-AGI-1 under the official evaluation protocol.

A natural hypothesis is that the identity embedding acts as a key for retrieving a task-specific ``program'' stored in the embedding space, while the small recursive trunk serves as a shared interpreter. Under this view, the system behaves like a neural hash map over puzzles. We emphasize that this interpretation is suggested by observed behavior rather than directly validated by probing internal representations.

This experiment does not imply label leakage in the usual sense: evaluation IDs do not overlap with training IDs, so the model is not simply memorizing training labels for test puzzles. It does, however, show that the ability to solve a particular evaluation puzzle is tightly linked to the associated ID token, which raises questions about how easily the model could be applied to genuinely new tasks without predefined identifiers.

\section{Experiment 3: Recursion Trajectory Analysis}

\subsection{Motivation}

TRM is designed to apply a fixed reasoning module recursively to a latent state. Training uses deep supervision across multiple recursion steps, suggesting that these steps are expected to contribute to improved performance. It is therefore natural to ask how much each recursion step contributes in practice and whether the model exhibits qualitatively different behavior as recursion depth increases.

\subsection{Design}

Using the verification checkpoint and the ARC-AGI-1 public evaluation set, we perform the following procedure:

\begin{enumerate}
    \item For each puzzle, we run the model's forward pass with recursion enabled up to the maximum number of latent cycles used during training (four in this checkpoint).
    \item After each recursion step ($t$), we decode the current latent state into an output grid and record the model's predicted solution.
    \item For each recursion depth ($t$), we compute Pass@1 by comparing the predictions at that depth to the ground-truth solution, using the full 1000-sample augmentation and majority-vote pipeline at that depth.
    \item As an extrapolation, we also evaluate performance when the recursion module is iterated beyond the nominal training depth, up to six steps.
\end{enumerate}

\subsection{Results}

As shown in Table~\ref{tab:trajectory}, the model reaches most of its final performance after the first recursion step: accuracy at step 1 is already very close to the accuracy at step 4. Subsequent steps yield smaller gains, and extending recursion beyond the training depth does not change accuracy.

\begin{table}[ht]
\centering
\caption{Pass@1 accuracy at different recursion depths (ARC-AGI-1, 400 tasks).}
\label{tab:trajectory}
\resizebox{\columnwidth}{!}{%
\begin{tabular}{lcc}
\toprule
Recursion step $t$ & Pass@1 & Relative to final (\%) \\
\midrule
1                  & 38.25\% & 94.4\% \\
2                  & 40.38\% & 99.7\% \\
3                  & 40.13\% & 99.1\% \\
4                  & 40.50\% & 100.0\% \\
6 (extrapolated)   & 40.50\% & 100.0\% \\
\bottomrule
\end{tabular}
}
\end{table}

\subsection{Interpretation}

For this checkpoint on ARC-AGI-1, most of the useful computation occurs in the first recursion step. The first application of the reasoning module determines the bulk of the outcome, and subsequent steps act as shallow refinements that provide modest improvements.

From a design perspective, recursion and deep supervision may still be important for learning a good initial mapping, since training encourages the model to improve predictions at multiple steps. However, at inference time the deployed model behaves more like a single-pass system with limited iterative correction than like a deeply multi-step reasoning engine. We do not claim that deeper recursion cannot be beneficial in other settings or for other checkpoints; our conclusion is restricted to the verification checkpoint and dataset studied here.

\section{Experiment 4: Training Dynamics Under Augmentation}

\subsection{Motivation}

The original TRM training setup employs heavy augmentation, generating many more training examples per puzzle than are present in the canonical dataset. This is made feasible by the model's high throughput. Augmentation can improve generalization by exposing the model to diverse inputs, but it can also complicate optimization and produce broad output distributions. We investigate these trade-offs by comparing early training behavior under canonical-only and heavily augmented regimes.

\subsection{Design}

We train two TRM models from scratch on the ARC-AGI-1 training set under the following regimes:

\begin{itemize}
    \item \textbf{aug 0 (canonical).} Only canonical task examples are used. No geometric or color augmentations are applied.
    \item \textbf{aug 1000 (heavy augmentation).} For each demonstration pair, 1000 random augmentations are generated using the same transforms as in the TRM pipeline.
\end{itemize}

Both models share identical architectures and optimization hyperparameters, except for the augmentation pipeline. To manage computational cost, we train each model for only the first 2,500 optimization steps out of a nominal schedule of approximately 778,000 steps (about 0.3\% of the full schedule). We treat this as a snapshot of early training behavior rather than as a fully converged model.

At this early checkpoint, we record:

\begin{itemize}
    \item Training loss and training accuracy on the respective training distributions.
    \item Evaluation Pass@1 and Pass@1000 on the ARC-AGI-1 public evaluation set by drawing 1000 independent augmented samples per puzzle and computing Pass@k directly over these sets, \emph{without} any majority-vote aggregation.
\end{itemize}

\subsection{Results}

As shown in Table~\ref{tab:trainmetrics}, both models reach high training accuracy on their respective training distributions, indicating substantial memorization even at this early stage. The canonical model trains slightly more easily, which is expected given its narrower distribution of inputs.

\begin{table}[ht]
\centering
\caption{Training metrics at step 2{,}500 ($\approx 0.3\%$ of full schedule).}
\label{tab:trainmetrics}
\resizebox{\columnwidth}{!}{%
\begin{tabular}{lcc}
\toprule
Metric & aug 0 (canonical) & aug 1000 (heavy) \\
\midrule
Train loss     & 0.3720  & 0.5987  \\
Train accuracy & 91.71\% & 88.98\% \\
\bottomrule
\end{tabular}
}
\end{table}

As shown in Table~\ref{tab:evalmetrics}, on ARC-AGI-1, the canonical model shows essentially identical Pass@1 and Pass@1000 at this step, whereas the heavily augmented model achieves slightly higher Pass@1000 than Pass@1. In the augmented regime, correct solutions appear among the sampled outputs more often than they appear as the top-ranked prediction.

\begin{table}[ht]
\centering
\caption{Evaluation metrics at step 2{,}500 (ARC-AGI-1, 400 tasks).}
\label{tab:evalmetrics}
\resizebox{\columnwidth}{!}{%
\begin{tabular}{lcc}
\toprule
Metric & aug 0 (canonical) & aug 1000 (heavy) \\
\midrule
Pass@1    & 0.38\% & 0.00\% \\
Pass@1000 & 0.38\% & 1.63\% \\
\bottomrule
\end{tabular}
}
\end{table}

\subsection{Interpretation}

Three qualitative observations emerge:

\begin{enumerate}
    \item Both models rapidly memorize substantial portions of their training data, but this does not immediately translate into strong evaluation performance on ARC-AGI-1.
    \item Under canonical training, when the model is correct, the correct solution tends to be its top prediction, leading to similar Pass@1 and Pass@1000. Under heavy augmentation, the model more frequently includes the correct solution among its samples, but its argmax prediction is often wrong at this early stage.
    \item Multi-sample metrics show benefits for the augmented model before single-pass accuracy improves, suggesting that heavy augmentation encourages a broader set of plausible outputs per puzzle.
\end{enumerate}

Because these models are extremely undertrained relative to the final TRM checkpoint, we treat these observations as qualitative. Fully trained models may have more concentrated solution distributions, but the early-stage behavior already illustrates how augmentation interacts with ensembling and coverage.

\section{Experiment 5: Efficiency and LLM Baseline}

\subsection{Motivation}

TRM's non-autoregressive design is explicitly motivated by efficiency considerations. A key claim of the original TRM paper is that the model's small size and parallel grid prediction enable heavy augmentation and ensembling that would be impractical for standard transformers. To provide concrete reference numbers, we compare TRM to a simple QLoRA fine-tune of Llama 3 8B on canonical ARC-AGI-1.

\subsection{Design}

We train a Llama 3 8B Instruct model \citep{grattafiori2024llama3} using QLoRA \citep{dettmers2023qlora} on the canonical ARC-AGI-1 training set. The setup is deliberately simple:

\begin{itemize}
    \item We use a standard QLoRA configuration with low-rank adapters.
    \item We train for a small number of epochs on the canonical examples only, without synthetic data.
    \item We do not use chain-of-thought prompting, test-time training, or specialized tokenization for ARC.
\end{itemize}

This baseline is intended as a low-resource, straightforward supervised fine-tuning regime. It is not meant to represent the state of the art in LLM-based ARC systems, which often rely on more elaborate training and inference strategies.

We then evaluate both TRM and the Llama baseline on ARC-AGI-1 under single-pass canonical inference and measure throughput, latency, and peak VRAM usage on an H100 GPU.

\subsection{Results}

Under this naive fine-tuning regime, the Llama 3 8B baseline achieves approximately 2.15\% Pass@1 on ARC-AGI-1, which indicates poor generalization from the small canonical dataset alone.

The efficiency profile on an H100 is summarized in Table~\ref{tab:efficiency}.

\begin{table}[ht]
\centering
\caption{Efficiency profile on NVIDIA H100 (80 GB).}
\label{tab:efficiency}
\resizebox{\columnwidth}{!}{%
\begin{tabular}{lccc}
\toprule
Model & Peak VRAM & Throughput & Latency \\
\midrule
TRM (7M parameters)      & 2.4 GB & 31.3 samples/s & 31.9 ms/sample \\
Llama 3 8B (QLoRA)       & 6.1 GB & 0.24 samples/s & 4170 ms/sample \\
\bottomrule
\end{tabular}
}
\end{table}

TRM requires substantially less memory and achieves much higher throughput than the Llama baseline on the same hardware, while also achieving higher accuracy in this setting.

\subsection{Interpretation}

These results confirm that TRM has a strong efficiency advantage over a generic 8B-parameter transformer in this setup and that its inductive bias is better aligned with ARC-style tasks under simple supervised training. However, efficiency alone does not fully explain TRM's performance. The ensemble and identity ablation experiments show that reported accuracy on ARC-AGI-1 depends on how this efficiency is used: TRM converts low latency into the ability to run extensive augmentation and voting, and it relies on puzzle-specific embeddings to organize behavior across tasks. The architecture, training procedure, and evaluation protocol are tightly coupled.

\section{Discussion}

This section synthesizes the empirical findings from our experiments and situates them within the broader context of recursive models and reasoning benchmarks. We distinguish direct observations from interpretive hypotheses and make clear where our conclusions are specific to the analyzed checkpoint.

\subsection{Puzzle Identity as a Central Conditioning Signal}

The puzzle-identity ablation shows that the TRM verification checkpoint is functionally dependent on the presence of the correct puzzle identifier. Replacing the puzzle ID with either a fixed ``blank'' token or a random token reduces accuracy to zero under the verification protocol, despite unchanged inputs and parameters. This implies that the model's ability to produce correct outputs for a given puzzle is tightly coupled to the corresponding identity embedding.

We interpret this as evidence that puzzle-ID embeddings play a central role in organizing the model's behavior. A natural behavioral hypothesis is that the identity embedding acts as a key into a space of task-specific behaviors, with the small recursive trunk serving as a shared computation mechanism. Under this view, a substantial fraction of the system's effective capacity resides in the embedding table and its interaction with the augmented training distribution, rather than solely in the 7M-parameter trunk.

This conclusion is specific to the studied checkpoint under the ARC-AGI-1 evaluation protocol. It does not imply that TRM-like architectures cannot be trained to rely less on explicit identity conditioning, nor that identity embeddings constitute label leakage in the usual sense.

\subsection{Shallow Effective Recursion}

The recursion trajectory analysis indicates that, for the verification checkpoint on ARC-AGI-1, most of the useful computation occurs in the first recursion step. The model reaches nearly all of its final accuracy at step 1, with subsequent steps providing modest refinements, and extending recursion beyond the training depth does not change performance.

These findings suggest that, although the model is trained with multiple supervised recursion steps, its deployed behavior is effectively shallow. The first application of the reasoning module appears to determine the bulk of the outcome, with later steps functioning mainly as minor corrections. Recursion and deep supervision may still be important for learning a robust initial mapping, but at inference time the model behaves more like a single-pass system with limited iterative refinement.

Again, this conclusion is restricted to the verification checkpoint and dataset studied here. Other TRM configurations or tasks might make more substantial use of deeper recursion.

\subsection{Augmentation, Ensembling, and the Shape of the Solution Distribution}

The combination of the ensemble analysis and the training dynamics experiments clarifies how augmentation and voting shape TRM's behavior. The 1000-sample augmentation and majority-vote pipeline contributes a significant absolute improvement in Pass@1 over single-pass canonical inference. In early training, a canonical-only model exhibits similar Pass@1 and Pass@1000, whereas a heavily augmented model shows higher Pass@1000 than Pass@1.

This pattern is consistent with a broader, more diffuse distribution of candidate outputs under heavy augmentation, where metrics that consider ``at least one correct sample'' can improve even when the argmax prediction remains wrong. Heavy augmentation encourages the model to represent a larger set of plausible outputs for each puzzle. In our training dynamics study, Pass@1000 is computed directly over raw samples without aggregation, so the improvement reflects increased coverage rather than better ranking. In the full verification pipeline, similar augmentation is combined with majority voting, which aggregates this broader distribution into a single prediction.

Because our training experiments are short-horizon, we view these observations as qualitative. Fully trained models may have more concentrated solution distributions, but the early-stage behavior already illustrates how augmentation and ensembling interact.

\subsection{Efficiency as an Enabler Rather Than a Standalone Explanation}

Our efficiency measurements show that TRM achieves much higher throughput and lower memory usage than a naive QLoRA fine-tuned Llama 3 8B baseline while also attaining substantially higher accuracy under a comparable canonical-only training regime. This confirms that TRM's architecture is not only efficient but also better aligned with ARC-style tasks than a generic transformer of similar training budget.

However, efficiency alone does not explain the observed performance. The ensemble and identity ablation experiments indicate that reported scores on ARC-AGI-1 depend on how this efficiency is used. The system converts low latency into the ability to run extensive augmentation and voting, and it relies on puzzle-specific embeddings to organize behavior across tasks. In other words, architecture, training procedure, and evaluation protocol jointly determine performance.

Our results therefore support a nuanced view: TRM's efficiency is a key enabling factor for heavy augmentation and ensembling, but the achieved accuracy on ARC-AGI-1 is the product of this efficiency interacting with strong per-task conditioning and shallow but effective recursion.

\section{Limitations}

We summarize the main limitations of our analysis to clarify the scope of our conclusions.

\begin{enumerate}
    \item \textbf{Single checkpoint and configuration.} All detailed ablations are performed on a single publicly released verification checkpoint with a specific recursion depth and training history. Other TRM checkpoints, or models trained with different hyperparameters or datasets, may exhibit qualitatively different behavior. Our claims should therefore be understood as applying to this particular instance of TRM, not to all possible TRM-like architectures.
    \item \textbf{Restricted datasets and tasks.} We focus exclusively on ARC-AGI-1 for both evaluation and most training experiments. We do not analyze ARC-AGI-2, Sudoku Extreme, Maze Hard, or other benchmarks where TRM has been reported to perform well \citep{jolicoeurmartineau2025trm}. Consequently, we cannot directly explain why the full TRM training pipeline achieves higher relative gains on some tasks or why performance differs between ARC-AGI-1 and ARC-AGI-2. Any connections to those results are conjectural.
    \item \textbf{Early training snapshot in augmentation experiments.} In our augmentation comparison, both models are trained for only a small fraction of the nominal schedule. The purpose of these experiments is to illustrate qualitative trends in training dynamics and solution distributions, not to characterize fully converged models. It is possible that the relative behavior of canonical-only and heavily augmented models changes later in training.
    \item \textbf{Naive large language model baseline.} The Llama 3 8B QLoRA baseline is intentionally simple. It uses a straightforward supervised fine-tuning setup on canonical ARC-AGI-1 examples, without synthetic data, task-specific tokenization, or advanced test-time training. As such, it does not represent the state of the art in LLM-based ARC systems. We use it only as a reference point for illustrating differences in inductive bias and efficiency, not as a definitive comparator.
    \item \textbf{Behavioral analysis only.} Our conclusions are based on input--output behavior (accuracy under different conditions, efficiency metrics) and limited code inspection. We do not perform a detailed analysis of internal representations or dynamics, for example via probing, attribution methods, or mechanistic interpretability tools. The ``neural hash map'' interpretation of puzzle-ID conditioning is therefore a hypothesis suggested by observed behavior, not a direct claim about the internal geometry of the embedding space.
    \item \textbf{Fixed evaluation protocol.} All experiments respect the ARC Prize verification pipeline's evaluation protocol, including specific augmentation schemes and scoring definitions. While this ensures comparability with reported verification scores, it also means that our analysis does not explore alternative inference strategies---such as different ensemble sizes, alternative voting schemes, or modified augmentation distributions---that might yield additional insight.
\end{enumerate}

\section{Reproducibility}

To facilitate further research, we provide a brief description of the resources needed to reproduce the experiments in this note.

\begin{itemize}
    \item \textbf{Code repository.} All scripts used for data preparation, evaluation, and experiments are contained in a public repository at \url{https://github.com/AntonioRoye/TinyRecursiveModels}, including configuration files, logging utilities, and experiment drivers.
    \item \textbf{Checkpoint retrieval.} The TRM verification checkpoint used in all ablations can be obtained from Hugging Face as \url{arcprize/trm_arc_prize_verification}. Scripts for downloading and converting this checkpoint into the format expected by the repository are provided.
    \item \textbf{Dataset preparation.} Scripts for preparing canonical and augmented ARC-AGI-1 datasets are provided. These scripts generate directories such as \texttt{data/arc-aug-0} and \texttt{data/arc-aug-1000} that correspond to the canonical and heavily augmented regimes, respectively.
    \item \textbf{Experiment scripts.} Dedicated entry points are provided for running the ensemble analysis, puzzle-identity ablation, recursion trajectory study, training dynamics comparison, and efficiency profiling. Each script reads configuration files that specify paths, hyperparameters, and random seeds.
\end{itemize}

We encourage readers to treat the reported numbers as reference values for this particular checkpoint rather than as definitive limits on what TRM-like architectures can achieve.

\section{Conclusion}

This technical note has presented a set of targeted empirical studies that clarify several aspects of the Tiny Recursive Model as instantiated in the ARC Prize verification checkpoint on ARC-AGI-1.

We have shown that test-time augmentation and majority-vote ensembling contribute a substantial absolute improvement in Pass@1 over single-pass canonical inference, that puzzle-identity embeddings are a strict dependency for this checkpoint under the verification protocol, and that the effective depth of recursion is shallow, with most performance achieved at the first recursion step. We have also shown that heavy augmentation modifies the structure of the model's predictions, increasing coverage under multi-sample metrics before improving single-pass precision, and that TRM's non-autoregressive design affords high throughput and low memory usage compared to a naive 8B LLM baseline.

Taken together, these findings suggest that TRM's success on ARC-AGI-1 is best understood as an interaction between a highly efficient architecture, strong per-task conditioning, and aggressive test-time compute, rather than as evidence of arbitrarily deep internal reasoning. Future work could extend this analysis by training and evaluating TRM variants with reduced or grouped puzzle-identity conditioning, systematically exploring different recursion depths at both training and inference time, and comparing TRM to alternative tiny architectures under matched augmentation and compute budgets.

We hope that this note serves as a useful reference for researchers studying recursive models on ARC and related benchmarks, and that it encourages more transparent reporting of the roles played by task identifiers, augmentation, and test-time compute in future work.

\bibliographystyle{plainnat}
\bibliography{paper}

\end{document}